%% file: main.tex
\documentclass[conference]{IEEEtran}
\usepackage{hyperref}
\IEEEoverridecommandlockouts
\usepackage{cite}
\usepackage{subcaption}
\usepackage{amsmath,amssymb,amsfonts}
\usepackage{algorithmic}
\usepackage{subcaption}
\usepackage{stfloats}
\usepackage{graphicx}
\usepackage{tikz}
\usepackage{amssymb}
\usetikzlibrary{arrows.meta, calc, decorations.pathreplacing}
\usepackage{textcomp}
\usepackage{xcolor}
\def\BibTeX{{\rm B\kern-.05em{\sc i\kern-.025em b}\kern-.08em
    T\kern-.1667em\lower.7ex\hbox{E}\kern-.125emX}}
\begin{document}

\title{Low-Overhead Error-Corrected QCNNs Using Bivariate Bicycle Codes}

\author{
\IEEEauthorblockN{Alejandro Rosales and Animesh Yadav}
\IEEEauthorblockA{School of Electrical Engineering and Computer Science \\
Ohio University, Athens, OH, USA \\
\{ar993823, yadava\}@ohio.edu}
}
\maketitle

\begin{abstract}
Quantum convolutional neural networks (QCNNs) combine the power of quantum computing and classical CNN for computational speedup in classification tasks. However, noise levels on state-of-the-art quantum devices remain too high for practical QCNN execution. In addition, despite the reliable surface code providing a method for error rates below a threshold value, they have a prohibitively large qubit cost. Recently introduced bivariate bicycle (BB) codes are of particular interest for their high error threshold, constant encoding rate, and linear code distance. Through simulation with realistic hardware noise sources, we demonstrate that a 4-qubit unprotected QCNN fails to converge and exhibits a worse learning rate compared to numerical simulations. Addressing both limitations, we propose a distance-4 BB quantum error-correction (QEC) technique for QCNNs. In doing so, we validate that our low-overhead QEC technique for QCNNS represents a step toward practical QCNNs.
\end{abstract}


\section{Introduction}
Advancements in machine learning (ML) have made it a crucial technology across various domains, from signal processing to healthcare, with applications such as speech recognition, computer vision, and drug discovery. However, ML procedures suffer from gradient vanishing in high-dimensional parameter spaces~\cite{Hochreiter1998, Pascanu2013} and quadratic sample sizes and training time for certain tasks~\cite{Huang2021}. On the other hand, quantum computing (QC) shows exponential computational speedups~\cite{LeCun2015, Nielsen2011, Hirvensalo2001}. The combination of these two technologies, termed quantum machine learning (QML), promises processing of data in exponentially large feature spaces~\cite{Biamonte2017, Havlicek2019} by leveraging quantum effects like superposition and entanglement, and more efficient navigation of high-dimensional optimization landscapes via quantum-enhanced optimization~\cite{McClean2018}. Although QML holds significant promise, the limited qubit counts of current noisy intermediate-scale quantum (NISQ) devices pose substantial challenges. Furthermore, these devices suffer from high physical error rates introduced by noise sources including stray electromagnetic fields, cosmic rays, and thermal and temporal decoherence of quantum states~\cite{Jay2017}.

Quantum error correction (QEC) is a method for protecting fragile quantum information while controlling qubits to induce a desired computation. For over two decades, the topological toric code has served as the canonical model for topological quantum error correction. \cite{Kitaev2003}. The code encodes two logical qubits into a $d \times d$ lattice of physical qubits, with the total qubit count scaling as $n = 2d^2$, where $d$ is the code distance. With minimum distance $d$, the code can correct up to $\left\lfloor \frac{d - 1}{2} \right\rfloor$ arbitrary single-qubit errors. The toric code thus provides reliable protection of quantum information, however it suffers from inefficient encoding rates that makes scaling to hundreds of logical qubits prohibitively resource-intensive. Recent advances in quantum Low-Density Parity-Check (qLDPC), most notably the bivariate bicycle (BB) code~\cite{Bravyi2024}, have emerged as a promising alternative. The constant encoding rate of BB codes enables fault-tolerant quantum memory with constant space overhead~\cite{Bravyi2024, Gottesman2014}, making them an attractive candidate for scalable architectures. However, a key open problem remains for the BB code is that it has yet to support fault-tolerant computation, particularly for deep-circuit algorithms and poses an obstacle to real-world QML use cases.

To this end, we introduce a constant-overhead QEC protocol that integrates with QML, particularly, quantum convolutional neural networks (QCNN). QCNNs are the quantum counterpart of classical convolutional neural networks (CNNs). These classical supervised learning architecture trains on labeled data to learn a mapping between inputs and their respective labels. QCNNs extend this framework by leveraging quantum properties such as quantum parallelism, to accelerate computation. However, practical QCNN execution on NISQ hardware is severely hindered by quantum decoherence and gate noise, which degrades the loss landscape and therefore compromise training and inference. The proposed constant-overhead QEC protocol directly mitigates these impairments, enabling more reliable and practical QCNN deployment on near-term quantum devices.

\subsection{Related Work}
In~\cite{Cong2019}, the authors introduced a QCNN architecture that incorporated the multi-scale entanglement renormalization ansatz (MERA) with QEC, demonstrating QCNNs for quantum phase recognition (QPR) of $1$D symmetry-protected topological (SPT) phases and designing a QEC scheme. Their QCNN uses $\mathcal{O}(\log N)$ variational parameters for input sizes of $N$ qubits. However, there is still a gap for QEC codes for error correction on deep circuits, like the QCNN, while being practical.

In~\cite{Bravyi2024}, the authors proposed the BB code that requires $n$ ancillary qubits, a depth-$7$ syndrome measurement circuit, and a degree-$6$ qubit connectivity graph, enabling efficient error correction with minimal overhead. Their code demonstrated reduced hardware requirements compared to the surface code and making fault-tolerant quantum memory feasible for near-term quantum processors. The application of the BB code for quantum computation is a high barrier, given the BB code works for Clifford gates only, unlike the Non-Clifford gates in the QCNN. In \cite{Wang2026}, the authors demonstrated low-overhead qLDPC codes, a distance-$3$ qLDPC code and a distance-$4$ BB code utilizing periodic execution of the syndrome extraction circuit. 

The work of~\cite{Bermejo2026} addresses the classical simulability and computational hardness of the functions of quantum models. The key distinction is that our research addresses trainability on physical hardware and how hardware noise introduces a stochastic, non-unitary perturbation at each circuit evaluation, degrading the fidelity of gradient estimates and making gradient estimates unreliable. This is orthogonal to questions of whether the learned function is classically hard to simulate or evaluate.

\subsection{Contributions}
Motivated by the the application of low-overhead BB code on a real $32$ long-range-coupled NISQ transmon circuit~\cite{Wang2026}, and the computational speed of machine learning training by the QCNN~\cite{Cong2019}, we apply the distance $4$ qLDPC codes for the QCNN. Specifically, we propose a distance-$4$ BB code \cite{Bravyi2024} combined with a Feed-Forward Neural Network (FFNN) as an error correction technique for the $4$-qubit QCNN \cite{Cong2019}. We integrate a constant encoding rate, linear code distance BB code that offers an error threshold of 0.3$\% $ for the standard circuit-based noise model to allow for low-overhead scaling to large QCNNs. Next, we compare the QCNN with the new QEC method against an unprotected 4 qubit QCNN in simulation at various NISQ error rates. We observed that the QCNN with the BB code achieves satisfactory results, which shows the implementation achieves a constant encoding rate and linear code distance offered by the BB code, with low qubit, online overhead for error correction. We further show that the QEC protocol can be sustained under low-overhead additional qubit resource requirement while reducing learning loss and improving convergence..

The rest of the paper is organized as follows. Section II presents the key background concepts described in our model implementation and the experiments conducted. Section III presents the proposed BB coded QCNN. Section IV present and discuss the results, comparing them to a plain QCNN architecture, and Section V concludes the paper along with some directions for future research.

\section{Background}

\begin{figure}[t]
    \centering
    \begin{subfigure}{0.49\linewidth}
        \centering
        \includegraphics[width=\linewidth]{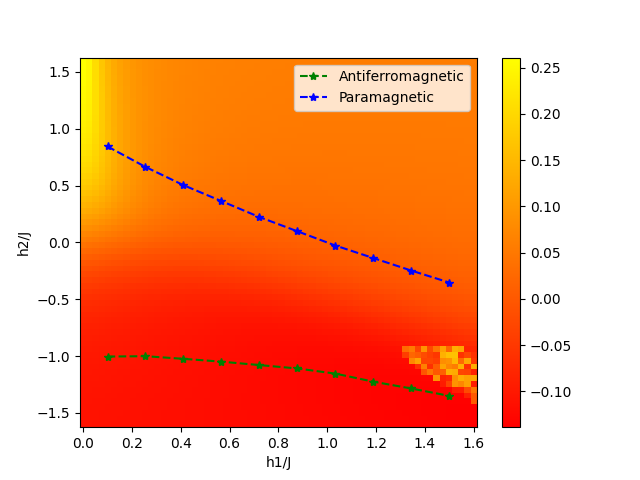}
        \caption{}
    \end{subfigure}
    \hfill
    \begin{subfigure}{0.49\linewidth}
        \centering
        \includegraphics[width=\linewidth]{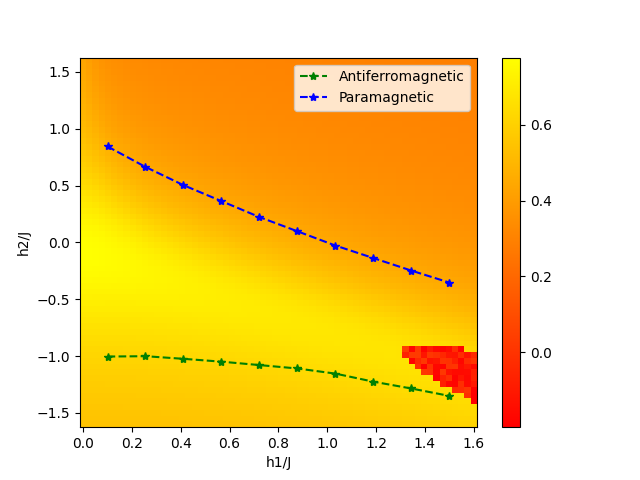}
        \caption{}
    \end{subfigure}
    \caption{(a) Points show a $64\times64$ test set of ground states over $h_1$ and $h_2$ for a Hamiltonian with $J=1$. Phase-boundary points (blue and red diamonds) come from infinite-size density-matrix renormalization group (DMRG). Colors indicate the circuit’s expectation values found via evolutionary search for $N=4$ spins using initial, untrained weights. (An artifact can be seen, possibly because of the low qubit spin state.) (b) Uses the same test set and with trained weights after 100 iterations.}
    \label{fig:placeholder}
\label{fig:init-vs-optimal-evaluation}
\end{figure}

\subsection{Quantum Convolutional Neural Network}

Classical CNNs has provided a successful ML framework for image recognition and are widely used to approximate functions due to their ability to capture hierarchical features \cite{LeCun2015, Bengio1995}. A QCNN is a quantum counterpart to the classical CNN that exploits quantum properties, such as quantum parallelism, to optimize computation. The QCNN architecture efficiently implements a hierarchical structure of quantum many-body states represented as an inverse MERA \cite{Cong2019}.

The MERA is a tensor network that is designed to efficiently represent quantum many-body states using a hierarchical and multi-layer structure. It can compute low energy properties of the system, described as the ground state $\left|\Psi_{\mathrm{GS}}\right\rangle \in \mathbb{C}^{\otimes N}\subseteq \mathcal{H}$, where $\mathcal{H}$ denotes the Hilbert space of a $N$-qubit system. 
Each layer $\tau$ of the MERA consists of disentanglers $U_\tau = \bigotimes_{i=1}^{N_\tau} u_i$ and isometries $W_\tau = \bigotimes_{i=1}^{N_\tau} w_i$, where $N_{\tau}$ denotes the number of disentanglers and isometries in that layer, $u_i$ denotes the parameterized two-qubit unitary operator applied at the $i$th convolutional unit of layer $\tau$, and $w_i$ denotes the corresponding isometric tensor at the reduced pooling subsystem. $N_{\tau}$ decreases with $\tau$ as sites are progressively coarse-grained. The transformation from a coarse-grained level to a finer-grained one is expressed as \[ V_\tau = U_\tau \circ W_\tau : \mathcal{H}_{\mathcal{L}_\tau} \longrightarrow \mathcal{H}_{\mathcal{L}_{\tau-1}}, \] where $\mathcal{H}_{\mathcal{L}_\tau} = \bigotimes_{s \in \mathcal{L}_\tau} \mathcal{H}_s$ is the Hilbert space of the coarse-grained lattice at layer $\tau$, with $\lvert \mathcal{L}_\tau \rvert < \lvert \mathcal{L}_{\tau-1} \rvert$ as sites are progressively reduced. The operator $\circ$ denotes function composition, and $V_\tau$ is the output unitary fed into the next layer~\cite{Pomarico2023}.

A QCNN can be thought of as a circuit-based ansatz of an inverse MERA with nested QEC. Learning occurs through gradual tuning of its initial unitary operations~\cite{Cong2019}. The QCNN acts on input states $|\psi_\alpha\rangle \in \mathcal{H}_{\mathcal{L}}$, where $\alpha = 1, \ldots, M$ indexes the training samples and $\mathcal{H}_{\mathcal{L}} \equiv \mathcal{H}_{\mathcal{L}_0}$ denotes the full input Hilbert space before any coarse-graining,
\begin{equation}
\mathcal{H}_{\mathcal{L}} = \bigotimes_{\ell \in \mathcal{L}} \mathcal{H}_\ell,
\end{equation}
and $\mathcal{H}_\ell \cong \mathbb{C}^2$ is the local qubit Hilbert space at site $\ell$.

The state of an individual site $\ell$ in any given step of the process, can be represented by its reduced density matrix $\rho^{[\ell]} = \operatorname{tr}_{\bar{\ell}}(|\Psi\rangle \langle \Psi|)$ and obtained by tracing out all other sites $\bar{\ell}$ except $\ell$. The target output for the QCNN is a realization of some state $|\psi_{\alpha}\rangle$, i.e., there is always a QCNN that recognizes an input state $|\psi_{\alpha}\rangle$ with some deterministic measurement outcome.

The QCNN training procedure is supervised approximation process defined over a training set $\mathcal{T}_{\text{train}}$ and a testing set $\mathcal{T}_{\text{test}}$, where a training sample $\vec{x} \in \mathcal{T}_{\text{train}}$ is provided to the algorithm together with its true label $m(\vec{x}) \in \{+1, -1\}$. Here, $m: \mathcal{T}_{\text{train}} \cup \mathcal{T}_{\text{test}} \rightarrow\left\{+1, -1\right\}$ is the underlying true map. The true labels of the test set $\mathcal{T}_{\text{test}}$ are not given to the algorithm, however, is given labeled training data $\left\{ (\vec{x},\, m(\vec{x})) \right\}_{\vec{x} \in \mathcal{T}_{\text{train}}}$. The QCNN learns an approximate map $\tilde{m}: \mathcal{T}_{\text{train}}\rightarrow\{+1,-1\}$ by minimizing a loss function assigns a penalty to deviations of $\tilde{m}(\vec{x})$ from $m(\vec{x})$ across the training set. Following training, the learned function $\tilde{m}$ is evaluated on unseen samples $\vec{x} \in \mathcal{T}_{\text{test}}$, where the goal is to minimize the generalization error $\mathcal{E} = \Pr_{\vec{x} \in \mathcal{T}_{\text{test}}} \left[ \tilde{m}(\vec{x}) \neq m(\vec{x}) \right]$. The training procedure will be discussed more in Section~\ref{sec:training-qcnn}

\subsection{Recognizing a 1D SPT Phase} \label{sec:qpr-for-1d-spt}

A QCNN takes advantage of the layered structure of the MERA to classify phase, where the input layer takes in an unknown $N$-input quantum state. After the input layer, the hidden layers are the convolutional layer, pooling layer, and fully connected layer \cite{Cong2019}, similarly to the CNN. The convolutional layer is given as a quasi-local unitary $U_\mathcal{L}$ and is applied in a translationally invariant manner for finite depth. During pooling, a portion of qubits is measured, and the outcome determines the unitary rotation $V_\mathcal{L}$ applied to nearby qubits, allowing for nonlinearity through the reduction of degrees of freedom. Unitary layers, such as unitaries in the convolutional and fully layers, apply quasi-local operations to existing qubits, while isometry layers, like those in the pooling layer, remove qubits in a state before applying unitary transformations on nearby states. The number of convolution and pooling layers remains fixed throughout, while the model learns the unitaries. The convolution and pooling layers are applied $d$ times. When the system is sufficiently small, a fully connected layer is applied on the remaining qubits as a unitary $F$. The outcome of the circuit is measured and the prediction is obtained.

The final operator measured by the circuit recognizes the SPT phase. The set of phases considered consists of ground states $\{|\psi_{\mathrm{GS}}\rangle\}$ corresponding to a family of Hamiltonians defined on an $N$-site spin-$\frac{1}{2}$ chain with open boundary conditions
\begin{equation}
    H = -J \sum_{i=1}^{N-2} Z_i X_{i+1} Z_{i+2} - h_1 \sum_{i=1}^{N} X_i 
    - h_2 \sum_{i=1}^{N-1} X_i X_{i+1},
    \label{eq:hamiltonian}
\end{equation}
where $X_i$ and $Z_i$ are Pauli operators acting on site $i$, $J > 0$ is the three-body interaction strength, $h_1$ is the transverse field strength, and $h_2$ is the nearest-neighbor coupling strength. The $\mathbb{Z}_2 \times \mathbb{Z}_2$ symmetry protection of the Haldane chain is generated by global $\pi$-rotations of every spin around the $X$ and $Y$ axes,
\begin{equation}
    R_x = \prod_j e^{i \pi \sigma_j^x} \quad \text{and} \quad R_y = \prod_j e^{i \pi \sigma_j^y},
    \label{eq:symmetry_generators}
\end{equation}
where $\sigma_j^x$ and $\sigma_j^y$ are the spin-$\frac{1}{2}$ operators at site $j$.

The one-parameter family of Hamiltonians considered for the Haldane phase, which is a SPT phase, is defined on a $1$-dimensional chain of $N$ spin-$1$ particles with open boundary conditions
\begin{equation}
H_{\text{Haldane}} = J_H \sum_{j=1}^N \mathbf{S}_j \cdot \mathbf{S}_{j+1} + \omega \sum_{j=1}^N \left(S_j^z\right)^2,
\label{eq:haldane}
\end{equation}
where $\mathbf{S}_j$ denotes the vector of spin-$1$ operators at lattice site $j$, $J_H > 0$ represents the exchange coupling strength, and $\omega \in \mathbb{R}$ is the single-ion anisotropy parameter.

The detection of the phase is as a measurement of the nonzero expected value $\langle Z_m \rangle $ middle qubit $m$ of the chain in the Z basis. Quantitatively, given a phase $|\psi_\alpha\rangle \in \mathcal{P}$, where $\mathcal{P}$ denotes the set of ground states belonging to a given phase, the expected value is computed directly from the state vector as
\begin{equation}
\langle Z_m \rangle = \sum_\ell \lambda_m \, |\langle \ell | \psi_\alpha \rangle|^2,
\end{equation}
where $m = \left\lfloor \frac{N+1}{2} \right\rfloor$, and the sum runs over all computational basis states $|\ell\rangle$, and $\lambda_m \in \{+1, -1\}^{N}$ is the eigenvalue of the Pauli-$Z$ operator on the middle qubit in the basis state $|\ell\rangle$.
\subsection{Stabilizer Codes}
QEC is crucial for protecting quantum information from background noise and gate errors that occur during processing. To achieve this, error correction protocols redundantly encode logical qubits into many physical qubits, some of which are ancilla qubits, to ensure errors can be detected and corrected. The ancilla qubits in a stabilizer code are repeatedly measured by parity check operators so that the wavefunction of the state of the logical qubits is not collapsed. The syndrome information is then used to detect and possibly correct errors that are believed to have occurred during processing. 

The number of errors that can be detected and corrected is dependent on the code. An $[[n,k,d]]$ QEC code encodes $k$ logical qubits into a QEC codeword of length $n$ to protect the system during quantum processing, where the code distance $d$ determines the minimum number of physical qubit errors needed to cause a logical error. A code of distance $d$ can correct up to $\left\lfloor \frac{d-1}{2} \right\rfloor$ errors and detect up to $d-1$ errors.

The information protected by the stabilizer code is defined over a stabilizer group $\mathcal{S}$, which is the Abelian subgroup of the Pauli group $P^{\otimes n}$. This group does not include $-I^{\otimes n}$ where $I$ is the $2 \times 2$ identity matrix in $P$. The stabilizer group $\mathcal{S}$ has $n-k$ independent generators $\mathcal{S}_i$, $i=1,...,n-k$. Each element in the stabilizer group $\mathcal{S}=\left\langle \mathcal{S}_1, \mathcal{S}_2, \ldots, \mathcal{S}_{n-k}\right\rangle$ is termed a stabilizer. The stabilizer generators perform distinct operations on the data qubits, however all stabilizer generators leave the encoded quantum state unchanged.

\subsection{Bivariate Bicycle Codes}
\label{sec:Bivariate Bicycle Codes}
The BB code is a type of Calderbank-Shor-Steane (CSS) stabilizer code that leverages bivariate polynomials over the quotient ring $R = \mathbb{F}_2[x,y]/(x^\ell-1, y^m-1)$, where $\ell$ and $m$ are positive integers, and $\mathbb{F}_2 = \{0,1\}$ is the binary field. An $[[n,k,d]]$ BB code $\operatorname{QC}(A,B)$ is a CSS code defined by two polynomials $A, B \in R$, where $\operatorname{QC}(A,B)$ denotes the associated quasi-cyclic code generated by the pair $(A,B)$, and has parameters
\begin{IEEEeqnarray}{rCl}
    n &=& 2\ell m, \\
    k &=& 2 \cdot \dim \left( \ker(\mathbf{H}^X) \cap \ker(\mathbf{H}^Z) \right), \\
    d &=& \min \left\{ |v| : v \in \ker(\mathbf{H}^X) \setminus \operatorname{rs}(\mathbf{H}^Z) \right\},
\end{IEEEeqnarray}
where $\mathbf{H}^X$ and $\mathbf{H}^Z$ are the parity-check matrices of the CSS code corresponding to the $X$-type and $Z$-type stabilizer checks, respectively, and $\mathbf{H}$ denotes a binary matrix. $\operatorname{ker}(\mathbf{H})$ denotes the set of vectors orthogonal to each row of $\mathbf{H}$, $\operatorname{rs}(\mathbf{H})$ is the linear span of its rows, and $|v| = \sum_{i=1}^n v_i$ is the Hamming weight of $v \in \mathbb{F}_2^n$.

\subsubsection{Encoding Logical Qubits}





The check matrices for a BB code denoted $\operatorname{QC}(A,B)$ with length $n=2\ell m$ is
\begin{equation}
\mathbf{H}^X = [A \mid B] \quad \text{and} \quad \mathbf{H}^Z = [B^T \mid A^T],
\label{eq}
\end{equation}
where the vertical bar indicates stacking matrices horizontally, and $T$ denotes matrix transpose. The bivariate polynomials are a pair of matrices 
\begin{equation}
A=A_1+A_2+A_3 \quad \text { and } \quad B=B_1+B_2+B_3,
\label{eq}
\end{equation}
such that each matrix $A_i$ and $B_j$ is a power of $x$ or $y$, while $\operatorname{dim} \mathbf{H}^X=\operatorname{dim} \mathbf{H}^Z=(n/2)\times 2$. Each row $v \in \mathbb{F}_2^n$ in $\mathbf{H}^X$ is a $X$-type operator, and similarly each row $v \in \mathbb{F}_2^n$ in $\mathbf{H}^Z$ is a $Z$-type operator, is defined as
\begin{equation}
X(v)=\prod_{j=1}^n X_j^{v_j} \quad \text { and } \quad Z(v)=\prod_{j=1}^n Z_j^{v_j},
\label{eq}
\end{equation}
respectively. The codes in Table 1 are described using the linear subspace associated with check matrices and polynomials $A$ and $B$ for high-rate, shown in \cite{Bravyi2024, Wang2026}.


\subsubsection{Syndrome Extraction}
The code standard $\operatorname{QC}(A,B)$ has a syndrome measurement (SM) circuit that continuously measures the syndrome of each check operator, which requires $n$ data qubits and $n$ ancillary check qubits to record the measured syndromes, for a total of $2n$ physical qubits. The SM circuit only applies $\text{CNOT}$s to pairs of qubits that are connected in the Tanner graph. 

The SM circuit starts and ends with a code dependent initialization and measurement cycle that determines the logical qubit initialization suited for the initial state and measuring logical qubits in a proper basis. The rest of the SM circuit is comprised of $N_c$ number of syndrome cycles (SC) repeated throughout the circuit periodically, where each SC measures syndromes for all $n$ check operators of the code.
\subsubsection{Decoding Algorithm}
Consider a $[[n,k,d]]$ BB code and an SM circuit $\mathcal{U}$ constructed with $N_c$ syndrome cycles defined in the SM section. Error correction is performed using a classical algorithm that takes a measured error syndrome as input and returns an estimate of the Pauli error that occurred during computation on the data qubits, accounting for all faults provided by the SM circuit. Note that the syndrome circuit itself may contain measurement errors. A proper guess occurs if the estimated Pauli error is a subset of the actual Pauli error up to a product computed via the check operators.



The SM step is followed by belief propagation with an ordered statistics postprocessing step decoder (BP-OSD)~\cite{Bravyi2024}. The unknown error for the linearized noise model used in the BB code is $\xi$, $D$ which denotes the decoding matrix, and $\sigma=\left[s^U | s^F\right]$ (where the vertical bar indicates stacking matrices horizontally) is the measured error syndrome. Using information sets ranked according to their reliability, BP-OSD finds an information set $I$ with the largest reliability. The output of BP-OSD is the solution of the system $D\xi=\sigma$ based on the most reliable information set $I$. This information set is used as the solution for the minimum weight error $\xi^*=\xi^*(s) \in\{0,1\}^N$  optimization problem. The guess for the unknown logical syndrome is given as
\begin{equation}
s^L=D^L \xi^*,
\end{equation}
where $L$ indexes the logical qubit degree-of-freedom. The BP-OSD decoder is applied separately to the decoding matrices $D_x$ and $D_z$, which are constructed from the parity-check matrices $\mathbf{H}^X$ and $\mathbf{H}^Z$, respectively. The result is a guessed $X$-type and $Z$-type errors $E_x$ and $E_z$, where the guessed final error is $E^*=E^*_xE^*_z$. This BP-OSD computes the upper bound for the code distance $d$. A large number of candidate BB codes with $n=\mathcal{O}(100)$ qubits is then searched that satisfy the criteria for the BB code given in the original paper \cite{Bravyi2024}.



\subsubsection{Error Correction}
After executing a quantum circuit the syndrome measurement outcomes are processed using BP-OSD. Using the final guessed X-type $E_x$ and Z-type error $E_z$ are used to correct the data qubits without measurement occurring.

\section{Proposed Bivariate-Bicycle Coded QCNN}

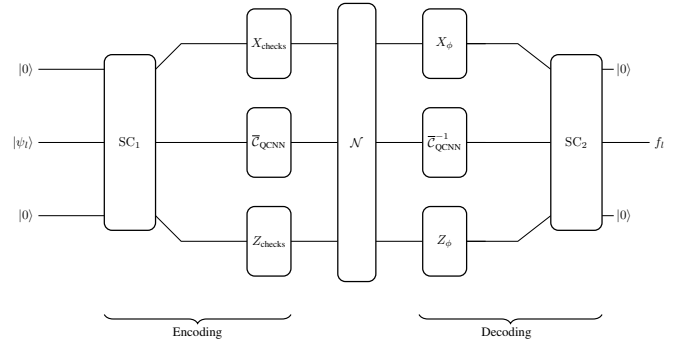
\begin{figure}[t]
  \centering
  \resizebox{\columnwidth}{!}{
    \input{fig_qec_circuit}}
  \caption{QEC-QCNN circuit.}
  \label{fig:fig_qec_circuit-circuit}
\end{figure}


In this section, we describe the parameters of the $4$-distance BB code and the $4$-qubit QCNN, followed by a description of how the QCNN is embedded within the BB code using transversal operations, ensuring that all QCNN operations remain within the protected code space. Finally, the Tanner graph topology of the trained network is used to correct errors in the QCNN.

\subsection{Training the QCNN} \label{sec:training-qcnn}
We used the  QCNN for quantum phase recognition (QPR) by applying it to a class of one-dimensional many-body systems. Specifically, the class considered is a $\mathbb{Z}_2 \times \mathbb{Z}_2$ phase. The ground states were numerically obtained using an infinite-size DMRG algorithm, following the method outlined in \cite{McCulloch2008} with a maximum bond dimension of $150$. The symmetry operators generating the $\mathbb{Z}_2 \times \mathbb{Z}_2$ symmetry are
\begin{equation}
    X_{\text{even}} = \prod_{i \in \text{even}} X_i 
    \quad \text{and} \quad 
    X_{\text{odd}} = \prod_{i \in \text{odd}} X_i,
\end{equation}
where ${X_i}$ is the Pauli-$X$ operator acting on site $i$. Each ground state energy density is obtained as a function of $h_2$ for fixed $h_1$. We then compute its second-order derivative to locate phase transitions. This approach to locating phase transitions via the second-order derivative of ground state energy densities follows \cite{Cong2019}.


As illustrated in Fig.~\ref{fig:init-vs-optimal-evaluation}, the QCNN circuit classifies whether the SPT phase exists as a $|S=1|$ Haldane chain or has transitioned to a $|S=1|$ paramagnetic phase or antiferromagnetic phase. When $\omega$ is zero or sufficiently small relative to $J$, the ground state, the phase belongs to the SPT phase. The phase is antiferromagnetic when the staggered field $h_1$ is sufficiently large relative to $J$, breaking the $\mathbb{Z}_2 \times \mathbb{Z}_2$ symmetry of the SPT phase. The critical point is identified as $h_2/J=0.423$ using infinite size DMRG numerical simulations.

During training, the untrained model is given a classified training set $\mathcal{T}_{\text{train}}=\{(|\psi_\alpha\rangle, y_{\alpha}): \alpha=1,...,M\}$, where $y_{\alpha}$  is the respective $0$ or $1$ binary classification for some input state $|\psi_{\alpha}\rangle$. The expected output of the QCNN is computed as $f_{\left\{U_i, V_i, F\right\}}\left(\left|\psi_\alpha\right\rangle\right)$ (The expected output computation will be discussed more in Section~\ref{sec:qec-for-qcnn}). The cost function of the model is computed using the mean-squared error (MSE)
\begin{equation}
\texttt{MSE}=\frac{1}{2 M} \sum_{\alpha=1}^M\left(y_i-f_{\left\{U_i, V_j, F\right\}}\left(\left|\psi_\alpha\right\rangle\right)\right)^2.
\label{MSEeq}
\end{equation}
The QCNN is constructed using isometric tensors in the pooling layer, where measuring a qubit lowers degrees of freedom for feature extraction \cite{Cong2019}. Since these measurement-based pooling operations are non-unitary and irreversible, standard quantum backpropagation cannot be applied through the full QCNN. An analytical gradient parameter-shift rule is used as an approximation scheme to compute the gradients, which provides unbiased gradient estimates for the variational unitary layers preceding the measurements.

The model learns by iteratively optimizing all initially assigned unitaries until convergence via a finite-difference scheme given the MSE cost function. To calculate the gradient matrix $M$ using the finite-difference scheme, the parameterized weights $\Theta$ in the QCNN circuit are perturbed by shifting $\theta_j \in \Theta$ by $\pm\epsilon$ for each parameter in the quantum circuit. The weights are then updated via gradient descent as
\begin{equation}
    \Theta \leftarrow \Theta - \eta \, M,
\end{equation}
where $\eta$ is the learning rate and $M$ is the gradient matrix of the MSE with respect to $\Theta$.

We evaluate phase recognition performance using 40 spin-$\frac{1}{2}$ ground states sampled along the $h_2$ axis ($h_1 = 0$), with QCNN weights trained from $0.616$ to $0.174$ after 100-iterations under ideal, noiseless statevector simulation. We will discuss the configuration of the simulation in Section \ref{sec:experimental-realization}. The trained model produces expectation values $\langle Z_m \rangle$ that decrease monotonically from approximately $0.94$ near $h_2 = 0$ toward $0.28$ at the far end of the paramagnetic region, yielding a smooth decision boundary consistent with the theoretical phase transition at $h_2 / J = 0.423$.

\subsection{Optimizing Quantum Error Correction}
\label{sec:Code Construction}
The QEC code is a BB code with parameters $[[18,4,4]]$, where $l = 3$ and $m = 3$, using $[a_1, a_2, a_3] = [1, 0, 1]$ and $[b_1, b_2, b_3] = [2, 0, 2]$. We derive the corresponding check matrices $\mathbf{H}^X=[A|B]$ and $\mathbf{H}^Z=[B^T|A^T]$ and compute the code parameters $k=4$ and $d=4$. The BB code circuit is then constructed in accordance with the architecture described in Section~\ref{sec:Bivariate Bicycle Codes}.

The feed-forward correction layer is trained and evaluated on stochastic Pauli noise injected into the $[[18,4,4]]$ BB code. Specifically, the four logical data qubits admit $2^4 = 16$ distinct computational basis states, such that the training set is 
\begin{equation} 
    \mathcal{\mathcal{F}_{\text{train}}} \;=\; \bigl\{\,y \in \{0,1\}^4\bigr\},
\end{equation} 
where each element $y = (y_1, y_2, y_3, y_4)$ encodes which of the four logical qubits has suffered a bit-flip error. At each SPSA iteration, every pattern $y \in \mathcal{\mathcal{F}_{\text{train}}}$ is prepared by applying $X$ gates to the mapped physical qubits. Concretely, the encoder uses $\zeta$  logical to map each basis state $\lvert y \rangle$, where $y \in \{0,1\}^4$, to a codeword $|\overline{y} \rangle \in (\mathbb{C}^2)^{\otimes 11}$ via transversal gate operations. Given a physical quantum circuit $\mathcal{C}$ acting on $4$ logical qubits, we construct a physically encoded circuit $\mathcal{C}_{\text{enc}}$ via the transversal gate operations. One classical register records the $11$ bits data measurement outcomes, respectively. using propagated through the full syndrome-extraction and feed-forward correction circuit, and decoded to yield a predicted logical label $\hat{y}$. 

Given the logical state space is finite and small the loss in Eq.~\eqref{eq:loss} is computed over all $N = 16$ patterns at every iteration, making each update equivalent to a full-batch gradient step over the complete input distribution. This exhaustive evaluation guarantees that the trained parameters $\boldsymbol{\phi}$ are assessed on every failure mode of the code, and that a loss of $0$ certifies perfect logical-state recovery across all possible single-layer error configurations.

To correct errors, we augment the syndrome extraction circuit with a parametrized feed-forward layer. This layer introduces $2m$ trainable rotation angles $\phi_i \in [0, 2\pi)$, where $R_y(\phi_i^X)$ and $R_y(\phi_i^Z)$ denote single-qubit $y$-axis rotation gates parameterized by the $i$th trainable angle corresponding to the $X$-check and $Z$-check ancilla qubits, respectively. These rotations are applied to the physical data qubits immediately following the syndrome vector, using controlled X and Z rotations.

The objective used during training are the logical state or logical labels. The set of true logical labels $\{y_i\}_{i=1}^{N}$ and the corresponding predicted logical labels $\{\hat{y}_i\}_{i=1}^{N}$ given in $\mathcal{F}_\text{train}$ returned by the decoder after circuit execution, the loss is given as
 
\begin{equation}
    \texttt{Loss}(\boldsymbol{\phi}) \;=\;
    \frac{1}{N}\sum_{i=1}^{N} \mathbf{1}\!\left[\hat{y}_i \neq y_i\right],
    \label{eq:loss}
\end{equation}
 
where $\boldsymbol{\phi}$ denotes the trainable parameters of the QEC feed-forward layer, distinct from the QCNN circuit parameters $\boldsymbol{\theta}$, and $\mathbf{1}[\cdot]$ denotes the indicator function. The loss therefore lies in $[0,1]$, with $0$ indicating perfect correction and $1$ indicating every frame is mislabeled. This normalization gives the SPSA gradient estimator a stationary, bounded signal that is independent of the number of test cases and the code distance.
 
The correction layer is embedded inside a quantum circuit evaluated by a noisy stochastic simulator. Since the loss function is a non-differentiable indicator function gradients of $\texttt{Loss}$ with respect to the trainable parameters $\boldsymbol{\phi}$ cannot be computed analytically~\cite{Spall1992}. SPSA is therefore employed as the optimizer, such that at each iteration $k$, the algorithm evaluates the a single syndrome cycle circuit $3$ times, regardless of the number of parameters $p$, making it practical for circuits with many syndrome checks.

Each training iteration proceeds in four stages.
\begin{enumerate}
    \item \textbf{Center evaluation.}
          The unperturbed circuit is evaluated to obtain the center-point loss $\texttt{Loss}_0 = \texttt{Loss}(\boldsymbol{\phi}^{(k)})$, where, $\boldsymbol{\phi}^{(k)}$ is weight vector at the $k$th iteration and is recorded for the learning-rate adaptation rule described below.
 
    \item \textbf{Positive perturbation.}
          A simultaneous perturbation vector $\boldsymbol{\Delta} \in \{-1,+1\}^p$ is drawn independently and uniformly at random. The parameters are shifted to $\boldsymbol{\phi}^{(k)} + c\,\boldsymbol{\Delta}$ where, $c = 0.2$\,rad is the perturbation magnitude, and the circuit is evaluated to obtain $\texttt{Loss}_+$.
 
    \item \textbf{Negative perturbation.}
          The parameters are shifted to $\boldsymbol{\phi}^{k} - c\,\boldsymbol{\Delta}$. Using the same realization of $\boldsymbol{\Delta}$, and the circuit is evaluated to obtain $\texttt{Loss}_-$.
 
    \item \textbf{Gradient step.}
          The SPSA gradient estimate for parameter $i$ is
 
          \begin{equation}
              \hat{g}_i \;=\;
              \frac{\texttt{Loss}_+ - \texttt{Loss}_-}{2\,c\,\Delta_i},
              \label{eq:spsa_grad}
          \end{equation}
 
          and a gradient descent step is applied
 
          \begin{equation}
              \phi_i^{k+1} \;=\;
              \left(\phi_i{k} - \alpha_k\,\hat{g}_i\right)
              \!\!\mod 2\pi,
              \label{eq:update}
          \end{equation}
 where $\alpha_k$ is the current learning rate. All parameters are updated simultaneously in a single step, with the base parameters restored from a snapshot taken before the perturbations were applied. Angles are wrapped modulo $2\pi$ throughout.
\end{enumerate}
 
The learning rate $\alpha_k$ is adapted at every step by a bold-driver rule~\cite{battiti1992}. After the gradient step, the center-point loss $\texttt{Loss}_0$ of the current iteration is compared against that of the previous iteration. If the loss has decreased, then the learning rate is increased by $1.05$, encouraging larger steps when progress is being made. If the loss has increased or remained constant, the learning rate is halved, contracting the search radius to
 
\begin{equation}
    \alpha_{k+1} \;=\;
    \begin{cases}
        1.05\,\alpha_k & \text{if } \texttt{Loss}_0^{(k)} < \texttt{Loss}_0^{(k-1)}, \\[4pt]
        0.5\,\alpha_k  & \text{otherwise.}
    \end{cases}
    \label{eq:bold_driver}
\end{equation}
 
All trainable parameters are initialized to zero, corresponding to identity rotations on every check qubit. Under this initialization the correction layer has no effect on the circuit output, so the very first center-point evaluation measures the uncorrected logical error rate. Subsequent iterations introduce progressively stronger corrections as SPSA explores the parameter space. A seeded random number generator is used for both the initialization procedure and the Bernoulli draws that produce the $\boldsymbol{\Delta}$ vectors, ensuring that training runs are exactly reproducible.

The decoding offline stage uses BB code decoding matrices to decodes the linearized noise model using BP-OSD \cite{Bravyi2024} by taking a measured error syndrome as input and outputting an estimated final error that occurred on the data qubits during computation.

\subsection{Error Correction for QCNN} \label{sec:qec-for-qcnn}

The QCNN suffers from the error-prone nature of quantum operations throughout the entirety of the circuit, so it naturally motivates the need for QEC. Our proposed QCNN architecture constructs the layers using transversal gate operations that conform to the Tanner graph structure of the BB code, denoted $\overline{\mathcal{C}}_{\text{QCNN}}$. Thus, allowing gates in each layer to preserve the code's locality and operations to occur in the protected subspace. The variational parameters are continuous and therefore implement non-Clifford rotations, which cannot be implemented transversally in the BB data block, allowing for errors to propagate through the circuit. 

To mitigate this, while avoiding the high overhead of magic state injection (MSI), we employ a FFNN as a syndrome-based neural decoder interleaved between each convolutional, pooling, and fully connected layer of the QCNN and the next. Thus, rather than correcting the non-Clifford rotations themselves, the FFNN intercepts the X-syndromes and Z-syndromes produced by the BB code's stabilizer checks after each noisy layer, and predicts per-qubit Pauli-X and Pauli-Z correction decisions before the next layer executes. This approach confines error propagation to within individual layers rather than allowing it to compound across the full circuit depth.

Concretely, this correction occurs between each hidden layer, prior to the next variational layer, so that the parameter updates computed by SPSA reflect gradient estimates over a cleaner logical state rather than one corrupted by accumulated gate noise, as would occur if correction were deferred to end-of-circuit. 

This process circumvents the classical post-processing bottleneck of BP-OSD, where belief propagation would need to complete on the order of 1$\mu s$~\cite{Acharya2024quantum, Caune2024realtime} for superconducting processors to keep pace with circuit execution. Current state-of-the-art decoders trade accuracy for latency to meet this tight decoding time budget~\cite{Viszlai2025gnn}, and BP-OSD does not meet this requirement at scale. Otherwise, physical qubits must sit idle between layers, during which $T_1$ (energy relaxation) and $T_2$ (dephasing) processes accumulate additional errors on the quantum hardware.

The FFNN's learnable weights are one per edge of the BB code's Tanner graph, such that the structure of the check and variable nodes respect the code's locality, mirroring the same geometric constraints imposed on the transversal gates. Its input is the X-syndrome bit string extracted from stabilizer CX measurements on the $[[18,4,4]]$ BB code's ancilla qubits, and its output is a binary correction vector indicating which physical data qubits require a corrective CX operation to return the logical state to the code space. Where, the $[[18,4,4]]$ BB code's stabilizer generators and logical operators are constructed following the procedure presented in \cite{Bravyi2024} (see section \ref{sec:Code Construction}).


The $4$-spin SPT input state is encoded into an $18$-physical-qubit state representing the logical state $|\psi_l\rangle$ of the $[[18,4,4]]$ code. The encoding is performed by mapping the four physical input spins to logical qubit degrees of freedom according to the sparse parity-check structure of the code. For each logical circuit $\mathcal{C}$, this procedure traverses every layer of the QCNN, including the input layer and final layer. The transversal operations then replaces each logical instruction with up to $|\mathcal{P}_c| \times |\mathcal{P}_t|$ physical gates. For a code with block size $n$ and a circuit of depth $d$ containing $g$ two-qubit gates, the encoded circuit has depth at most $\mathcal{O}(d)$ and gate count at most $\mathcal{O}(g \cdot n^2)$.
Specifically, each gate $G\in\mathcal{C}$ acting on logical qubit $q_i$ is replaced by the transversal operation
\begin{equation}
    G^{\otimes |\mathcal{P}_i|} = \bigotimes_{j \in \mathcal{P}_i} G_j,
\end{equation}
where $\mathcal{P}_i \subseteq \{0,1,\ldots,n-1\}$ is the set of physical qubit indices corresponding to the $i$th logical qubit under the code's qubit mapping $\zeta: q_i \mapsto \mathcal{P}_i$.

A logical CNOT between control qubit $q_c$ and target qubit $q_t$ is implemented transversally as
\begin{equation}
    \mathrm{CNOT}^{\otimes} = \bigotimes_{\substack{j \in \mathcal{P}_c \\ k \in \mathcal{P}_t,\; j \neq k}} \mathrm{CNOT}_{j \to k},
\end{equation}
applying a physical CNOT from each control physical qubit to each target physical qubit, excluding self-connections. Gates involving stabilizer check qubits or reset operations are passed through without transversal expansion, as these are already defined at the physical level in the encoded circuit.

The $\overline{\mathcal{C}}_{\text{QCNN}}$ then acts on the logical state, which undergoes noise $\mathcal{N}$ as.
\begin{equation}
\label{eq:bb-qcnn-function}
    f_l = \sum_{\lvert\psi_l\rangle \in \{\lvert{\pm x, y, z}\rangle\}} \langle\psi_l\rvert\, \mathcal{M}_q^{-1}\!\left( \mathcal{N}\!\left( \mathcal{M}_q\!\left( \lvert\psi_l\rangle\langle\psi_l\rvert \right) \right) \right) \lvert\psi_l\rangle,
\end{equation}
where $\mathcal{M}_q$ $(\mathcal{M}_q^{-1})$ denotes the encoding (decoding) map of the BB code, $\lvert{\pm x, y, z}\rangle$ are the $\pm 1$ eigenstates of the Pauli $X$, $Y$, $Z$ operators, and $f_l$ is the logical output of
\begin{equation}
\overline{\mathcal{C}}_{\mathrm{QCNN},\,\{U_i,\, V_j,\, F\}}
\!\left(\lvert\psi_l\rangle\right).
\end{equation}

The initial physical system two 9-qubit data registers $\mathcal{D}_L$ and $\mathcal{D}_R$, one 9-qubit $X$-check ancilla register $\mathcal{A}_X$, and one 9-qubit $Z$-check ancilla register $\mathcal{A}_Z$. However, the transversal qubit mapping has two 9-qubit data registers $\mathcal{D}_L$ and $\mathcal{D}_R$ which are redundant. The four logical qubits map onto only 11 distinct physical qubit indices $\{0, \ldots, 10\}$, with qubits $\{0,2,5,6\}$ shared across multiple logical qubits and the remaining indices each carrying information unique to a single logical qubit. Consequently, the two data registers collapse into a single 11-qubit register $\mathcal{D}$, reducing the full system from 36 to 29 qubits without loss of information.


The encoder uses $\zeta$  to map each basis state $\lvert i \rangle$, where $i \in \{0,1\}^4$, to a codeword $\lvert \overline{i} \rangle \in (\mathbb{C}^2)^{\otimes 11}$ via transversal gate operations. The evolved circuit is a Hilbert space of dimension $2^{n_\text{total}}$, where $n_\text{total} = n_\text{left} + d_\text{data} + n_\text{right} = 29$. Given $9$ X-check ancillas, $11$ data qubits, and $9$ Z-check ancillas. Given the ancilla qubits are initialized to $|0\rangle$ and, in the noiseless statevector simulation, remain unentangled with the data register, the data-qubit marginal state can be extracted without a partial trace. The reduced statevector is then decoded into the logical basis by iterating over all $2^{d_\text{data}}$ computational basis states of the data register. The corresponding logical state is then determined by
\begin{equation}
\mathbf{v} \;=\;
    \left[\,\bigl(L_X\,\mathbf{b}\bigr) \bmod 2\,\right],
    \label{eq:logical_decode}
\end{equation}
The complex amplitude $\alpha_j$ is then accumulated into the entry of the logical statevector $|\psi_l\rangle \in \mathbb{C}^{2^N}$ indexed by $v = \sum_{k} v_k \, 2^{N-1-k}$, so that
 
\begin{equation}
    |\psi_l\rangle \;=\;
    \sum_{j=0}^{2^{d_\text{phys}}-1}
    \alpha_j \,
    \bigl|\mathbf{v}(\mathbf{b}_j)\bigr\rangle.
    \label{eq:logical_sv}
\end{equation}
The resulting $2^{N}$-dimensional statevector is passed to the prediction stage, where the expectation value of the middle-qubit observable is evaluated to produce the final classification output.

\section{Experimental Results and Discussion} \label{sec:experimental-realization}
\subsection{Simulation Setup} We let each gate type is subject to an independent error probability $p$, with errors injected as single- or two-qubit Pauli operators drawn uniformly from $\{I, X, Y, Z\}$. The two-qubit CNOT gates are assigned errors from the full 15-element two-qubit Pauli group $\{I, X, Y, Z\}^{\otimes 2} \setminus \{II\}$, with each error type drawn uniformly given that an error occurs with probability $p$. Idle qubit errors are inserted to maintain a consistent time boundary across all data qubits. Idle qubits are modeled per layer, such that any qubit not involved in an active gate receives a uniformly random Pauli error $\{I, X, Y, Z\}$ with probability $p$, independently per qubit. Measurement operations are subject to a pre-measurement bit-flip, phase-flip, or $Y$-error at the same rate. Further, we implement a stochastic Pauli noise model applied at the circuit level \cite{Bravyi2024}. The considered noise model configuration is used to sweep over physical error rates $p \in \{0.0001, 0.001, 0.003\}$, which span the near-threshold regime relevant to NISQ-era demonstrations. This noise model approximates a symmetric depolarizing channel on each gate type and serves as a baseline for evaluating the robustness of our BB-code-protected QCNN against gate-level noise. All noise simulations used the same seeded random number generator for reproducibility.

\subsection{Training and Evaluation}
The 4-qubit QCNN was first trained under ideal statevector simulation using a dataset of 40 labeled ground states evenly spaced along the line $h_2 = 0$, where the Hamiltonian is exactly solvable via the Jordan-Wigner transformation and labeled $\pm 1$~\cite{Sachdev2011}. Training ran for 100 iterations using the finite-difference gradient scheme described in Section~II-B.

Fig.~\ref{fig:init-vs-optimal-evaluation} plots and compares statevector training training loss for $4$-qubit QCNN and 11-qubit transversal QCNN under error rates $p \in\{$ 0.0$\%$, 0.01$\%$, 0.1$\%$, 0.3$\%\}$. 

    \begin{figure}[h]
        \centering
        \includegraphics[width=\linewidth, height=5cm]{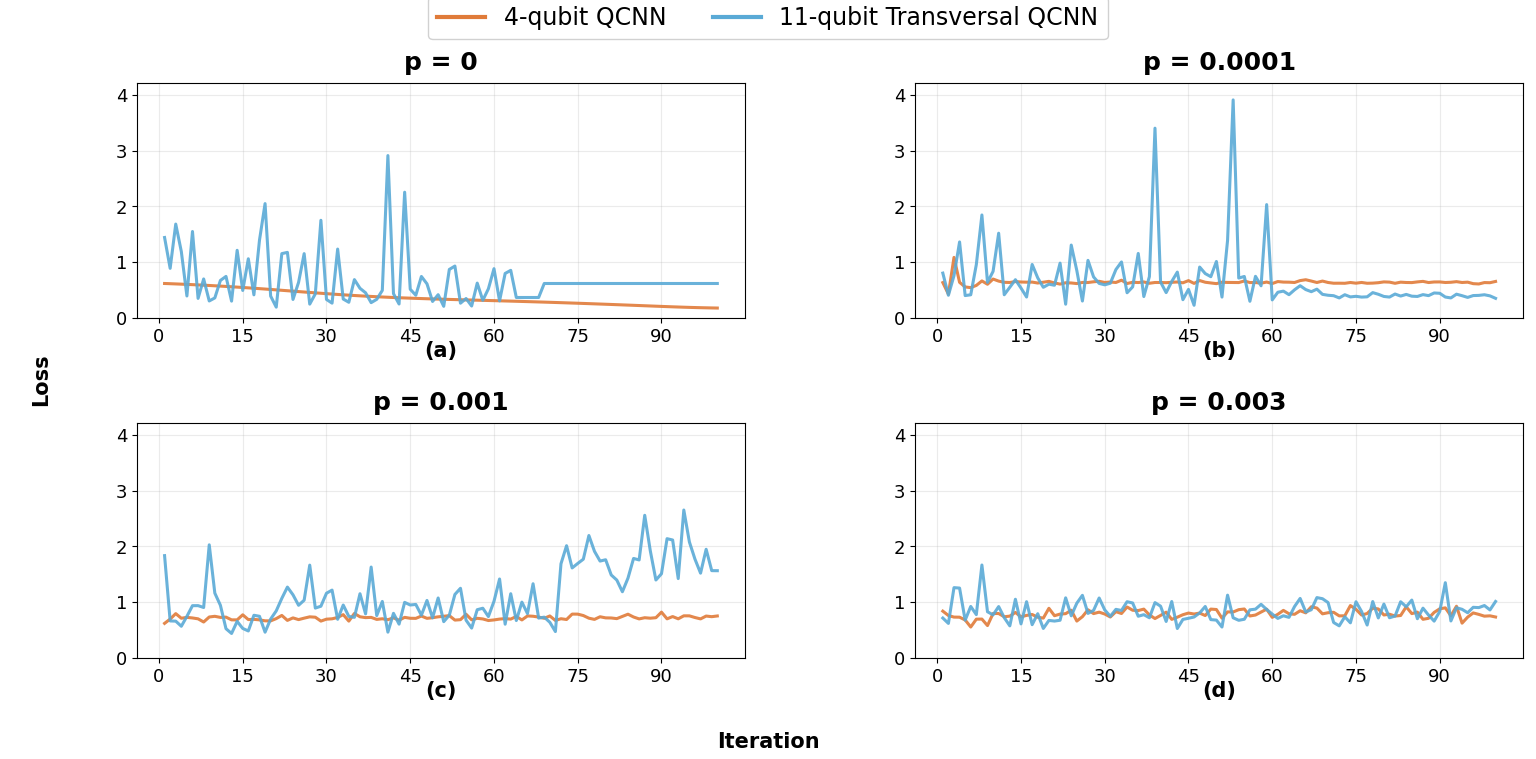}
        \caption{Statevector training loss vs iteration for $4$-qubit QCNN and 11-qubit transversal QCNN.}
        \label{fig:init-vs-optimal-evaluation}
    \end{figure}
    


In Fig.~\ref{fig:init-vs-optimal-evaluation}a, under noiseless (i.e., $p=0$) statevector training, the standard QCNN converges smoothly and monotonically from $0.616$ at to $0.174$ and reached a stable plateau by iteration 60. In comparison the transverse QCNN started at $1.442$ and exhibits substantial loss variance during the first 55 iterations, with a quick peak exceeding 2.0 and recurring peaks near iterations 20 and 45. After iteration 55, the transversal circuit stabilizes and plateaus at $0.616$, with the lowest loss at iteration 19 of $0.19$ and oscillating around losses of about $0.25$ from iterations 20 to 60. This variability and performance gap of the transversal QCNN under noiseless conditions suggest structural constraints due to the transversal gate set. We believe that the application of identical unitary operation fault-tolerantly across all physical reduces, the effective expressibility of the ansatz. Updates made to the weights are magnified by the repetition of the gates across many physical qubits. While this property is designed to minimize physical error rates on a logical qubit, it also causes large shifts for the finite-difference scheme when traversing the gradient. This can be seen by large steps taken during early-training, before plateauing after the learning loss was lowered, due to fine tuning from the bold driver.
 
In Fig.~\ref{fig:init-vs-optimal-evaluation}b, for $p=0.001$, the standard QCNN fails to learn and converge entirely. The loss starts at $0.632$ and after 100 iterations is at $0.652$. The QCNN reaches its lowest loss of $0.42$ at iteration 2, after which it never drops below a loss of $0.6$. The transversal QCNN starts at $0.802$ oscillates early on, however, start learning and descending around iteration 45, ending at $0.348$. The lowest loss recorded for the transversal QCNN was $0.224$ at iteration 46. The transversal QCNN is noise-resilient by construction. Given the transversal QCNN learned better than the standard QCNN without noise, we believe the error corrected QCNN will train better, however, with limited hardware, we leave simulating 29 qubits on classical hardware for future research.

In Fig.~\ref{fig:init-vs-optimal-evaluation}c, for $p = 0.003$, the highest noise level tested for training. The standard QCNN begins at a loss of $0.714$ at iteration~1 and never improves in any sustained amount. We assume this is because the bold-driver schedule triggers repeated halving events from as early as iteration~3, collapsing the learning rate from $10^5$ to below 
$10^{-8}$ by iteration~100, at which point the loss reads $1.565$ to a loss higher than the starting loss.

In Fig.~\ref{fig:init-vs-optimal-evaluation}d, for $p = 0.003$, the transversal QCNN  follows a similar loss volatility but exhibits markedly higher loss magnitudes throughout. The initial loss started at $0.714$ in iteration~1, the model oscillates chaotically across the full 100  iterations, with losses regularly exceeding $1.0$ and peaking above $1.6$ on multiple occasions at iterations 8 and 38, respectively. The learning rate undergoes the same rapid collapse as the standard variant, falling below $10^{-8}$ before iteration~100, and the final loss of $1.565$ provides no improvement over initialization. Neither architecture recovers a downward trend after the learning rate saturates near zero, 
and neither approaches the $0.348$ final loss achieved by the transversal QCNN at $p = 0.001$. This shows that an error rate of $p = 0.003$ is too high for trainability of both the standard and transversal QCNN under the present finite-difference scheme configuration. The noise-resilience advantage of the transversal design observed at $p = 0.001$ does not extend to this error rate.

\begin{figure}[h]
        \centering
        \includegraphics[width=\linewidth]{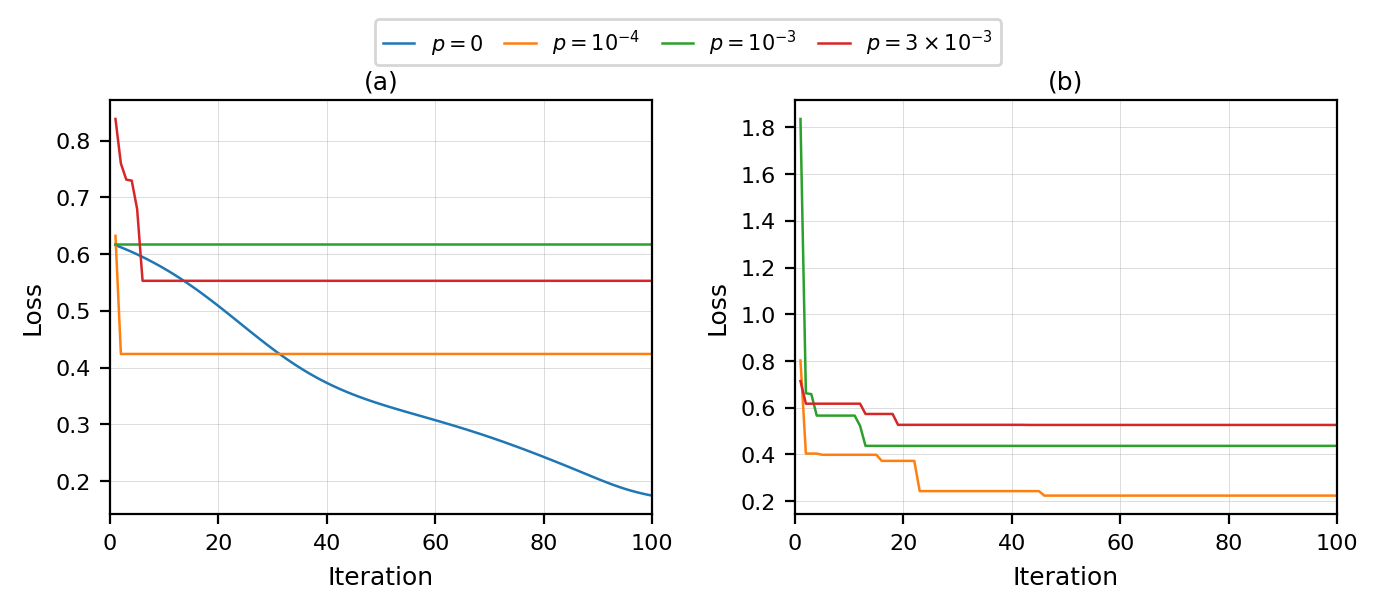}
        \caption{Training loss versus iteration (a) the 4-qubit QCNN and (b) the 11-qubit transversal QCNN.}
        \label{fig:qcnn-v-trans-min}    
\end{figure}
 
Fig.~\ref{fig:qcnn-v-trans-min} plots the best-so-far training loss over 100 iterations for the standard QCNN (a) and transversal QCNN (b) under different NISQ noise levels $p \in \{0, 0.0001, 0.001, 0.003\}$. Each plotted value reflects the lowest loss achieved up to that iteration rather than the raw per-iteration loss. Under noiseless conditions, the standard QCNN converges smoothly and monotonically to a final loss of approximately $0.174$, whereas all noisy variants stall early. At $p=0.001$ and $p=0.003$ plateauing immediately near $0.62$ and $0.57$ respectively, and $p=0.0001$ halting around $0.43$ after a few iterations. All four noise levels for the transversal QCNN descend rapidly within the first 10 iterations and continue improving, with the noiseless curve reaching approximately $0.21$, $p=0.0001$ settling near $0.22$, and even the noisiest conditions ($p=0.001$ and $p=0.003$) converging to losses of roughly $0.43$ and $0.50$ respectively. This contrast demonstrates that the transversal architecture retains trainability at noise levels that completely suppress learning in the standard QCNN, confirming its resilience at NISQ-era error rates.

\begin{figure}[h]
        \centering
        \includegraphics[width=\linewidth]{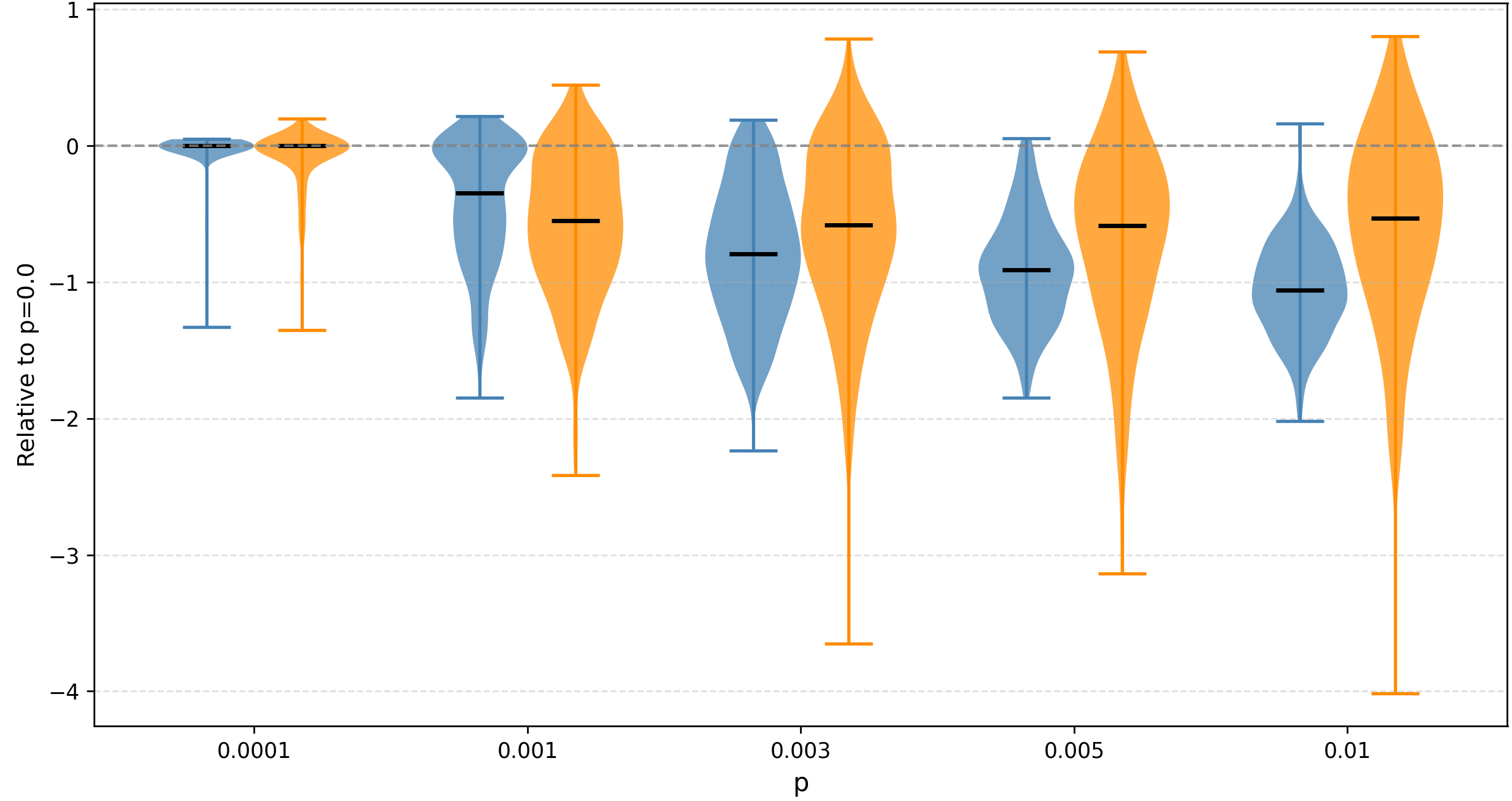}
        \caption{The violin plots of QCNN predictions (blue) vs Transversal QCNN (orange). Values are normalized to the noise-free median.}
        \label{fig:violin}
    \end{figure}    
Fig.~\ref{fig:violin} shows violin plots of qubit prediction values for the standard QCNN (in blue) and logical prediction values for the transversal QCNN (in orange) across Pauli noise levels normalized to the noise-free statevector predictions. Each noise level has $n=50$ predictions over 25 random seeds. At $p=0.0001$, both architectures remain tightly concentrated near zero, indicating minimal deviation from the noiseless baseline, as noise increases to $p>=0.001$, both distributions shift downward and broaden, reflecting degraded and increasingly variable predictions. The standard QCNN consistently produces a more compact and symmetric distribution with its median tracking closer to zero at lower noise levels, reaching approximately $-1.05$ at $p=0.005$ and $-1.2$ at $p=0.01$. The transversal QCNN, by contrast, exhibits wider distributions with heavier tails extending well below $-2$ at higher noise levels, and while its median remains slightly higher than the standard QCNN at $-0.7$ and $-0.6$ for $p=0.005$ and $p=0.01$ respectively, this comes with substantially greater variance and outliers. It can be concluded that the transversal gate structure preserves higher median prediction quality at elevated noise levels at the cost of greater variance, and that this preservation of median predictions may allow the optimizer to identify clearer local minima during noisy training.

\begin{figure}[h]
        \centering
        \includegraphics[width=\linewidth]{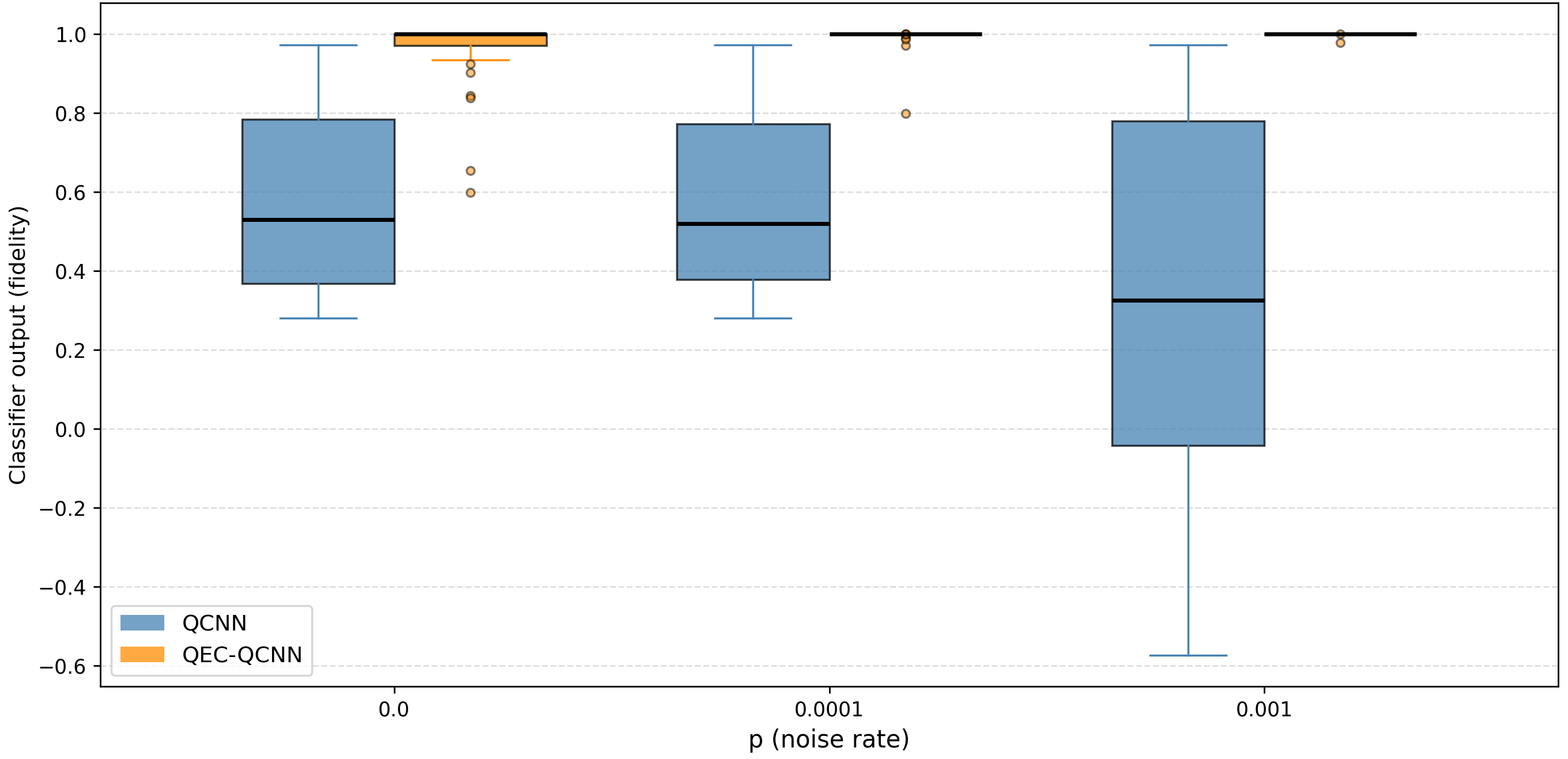}
        \caption{$4$-qubit QCNN prediction of an antiferromagnetic and paramagnetism phase with the trained model. The QCNN (blue) and the QEC-QCNN (orange) are shown.}
        \label{fig:box-wisker}
\end{figure}
The QEC-QCNN ~\ref{fig:box-wisker} shows that the FFNN trained on Clifford gates over corrects for errors, and Non-Clifford Gates are not apart of the training data, and appear to the FFNN soft decoder as errors. This causes bit and phase flips to the entangled data, causing predictions to entropy to predictions of 1. To improve results, the FFNN soft decoder should be trained with the QCNN model, for noise-aware error correction, or trained on a pre-trained model. Thus, errors in application match errors seen during training.

\subsection{Qubit Resource Analysis}
The toric surface code encodes logical qubits into a physical qubit lattice with overhead scaling as $n \propto d^2$, achieving an error threshold of approximately $0.67\%$-$0.81\%$ under circuit-level depolarizing noise using gauge-fixing techniques~\cite{Higgott2021}, below which increasing $d$ exponentially suppresses logical errors. However, encoding $k=4$ logical qubits at distance $d=4$ requires roughly $64$ physical qubits plus $400$--$1{,}000$ additional qubits per magic state distillation factory~\cite{Litinski2019,EastinKnill2009} to supply the non-Clifford gates essential to QCNN variational layers at current gate error rates~\cite{IBMHeronR2_2024}, placing total physical qubit requirements between $10^4$ and $10^6$~\cite{GidneyEkera2021,Lee2021,Babbush2021}. This is a prohibitive overhead for current hardware. By contrast, the $[[18,4,4]]$ BB code encodes the same $4$ logical qubits into $29$ physical qubits via a shared-index transversal mapping~\ref{sec:qec-for-qcnn}, with non-Clifford corrections handled by the interleaved FFNN decoder, eliminating factory overhead at the cost of a lower error threshold ($\sim 0.3\%$). This makes it far better suited to near-term operation.

\section{Conclusion and Future Directions}




In this study, we considered the problem of training instability of QCNN observed under NISQ hardware noise. We proposed the integration of BB code to overcome this issue. The $[[18,4,4]]$ BB code provides a mapping for physical to logical qubits and spins via the sparse parity check matrices to transversally encode a 4-qubit QCNN into an 11-qubit QCNN logical structure. The transversal QCNN, by contrast, retains meaningful learning progress at $p=0.001$, and converges to a final loss of 0.348, demonstrating that fault-tolerant encoding confers measurable noise resilience even without a fully trained decoder. At higher noise rates, such as $p=0.003$, neither architecture recovers. This indicates that the gradient landscape is too difficult for current QCNN models without meaningful error correction. These findings necessitate the joint error mitigation and error correction between the FFNN soft decoder and the transversal QCNN. When higher qubit counts become available, qLDPC codes with larger distances and more logical qubits can be applied to deeper QCNN models. Further, by co-training the decoder alongside the QCNN variational parameters, or by initializing from a pre-trained QCNN model, the correction layer can be exposed to the same error distribution encountered during inference, aligning the noise seen at training time with that seen in application.

\section*{Acknowledgment}
The authors would like to thank Jorma Kilpi and Olli Apilo of VTT 
Technical Research Centre of Finland for their insightful feedback to this work.

\bibliography{quantum_ref.bib}
\bibliographystyle{IEEEtran}


\end{document}

%% file: fig_qec_circuit.tex
\begin{tikzpicture}[>=Stealth, line width=0.8pt, font=\normalsize]
 
\newcommand{\measbox}[2]{%
  \draw[rounded corners=3pt, line width=0.7pt]
    (#1-0.28,#2-0.25) rectangle (#1+0.28,#2+0.25);
  \draw[line width=0.5pt] (#1-0.16,#2-0.08) arc(180:0:0.16);
  \draw[->,line width=0.5pt] (#1,#2-0.08) -- (#1+0.18,#2+0.16);
}
 
\newcommand{\gbox}[5]{%
  \draw[rounded corners=6pt, line width=1.0pt, fill=white]
    (#1-#3,#2-#4) rectangle (#1+#3,#2+#4);
  \node at (#1,#2) {#5};
}
 
%
%
%
%
 
\draw (-2.50, 2.00) -- (-0.70, 2.00);
\draw (-2.50, 0.00) -- (-0.70, 0.00);
\draw (-2.50,-2.00) -- (-0.70,-2.00);
 
\draw (0.70, 2.00) -- (1.40, 2.70) -- (3.20, 2.70);
\draw (0.70, 0.00) -- (3.20, 0.00);
\draw (0.70,-2.00) -- (1.40,-2.70) -- (3.20,-2.70);
 
 
\foreach \yw in {2.70, 0.00, -2.70}{
  \draw (4.40,\yw) -- (5.68,\yw);
}
 
\foreach \yw in {2.70, 0.00, -2.70}{
  \draw (6.72,\yw) -- (8.00,\yw);
}
 
\foreach \yw in {2.70, 0.00, -2.70}{
  \draw (9.20,\yw) -- (9.72,\yw);
}
 
\draw (9.20, 2.70) -- (10.60, 2.70) -- (11.50, 2.00);
\draw (9.20, 0.00) -- (11.50, 0.00);
\draw (9.20,-2.70) -- (10.60,-2.70) -- (11.50,-2.00);
 
\draw (12.90, 0.00) -- (14.20, 0.00);
\draw (12.90, 2.00) -- (13.22, 2.00);
\draw (12.90,-2.00) -- (13.22,-2.00);
 
 
\gbox{0.00}{0.00}{0.70}{2.40}{$\mathrm{SC_1}$}
 
\gbox{3.80}{2.70}{0.60}{0.95}{$X_\text{checks}$}
\gbox{3.80}{0.00}{0.60}{0.95}{$\overline{\mathcal{C}}_{\text{QCNN}}$}
\gbox{3.80}{-2.70}{0.60}{0.95}{$Z_\text{checks}$}
\gbox{6.20}{0.00}{0.52}{3.80}{$\mathcal{N}$}
 
\gbox{8.60}{2.70}{0.60}{0.95}{$X_\phi$}
\gbox{8.60}{0.00}{0.60}{0.95}{$\overline{\mathcal{C}}_{\text{QCNN}}^{-1}$}
\gbox{8.60}{-2.70}{0.60}{0.95}{$Z_\phi$}
 
\gbox{12.20}{0.00}{0.70}{2.40}{$\mathrm{SC_2}$}
 
 

\node at (13.50, 2.00) {$|0\rangle$};
\node at (13.50, -2.00) {$|0\rangle$};
 
 
\node[anchor=east] at (-2.50, 2.00) {$|0\rangle$};
\node[anchor=east] at (-2.50, 0.00) {$|\psi_l\rangle$};
\node[anchor=east] at (-2.50,-2.00) {$|0\rangle$};
 
 
\node[anchor=west] at (14.20, 0.00) {$f_{l}$}; 
 
\draw[decorate, decoration={brace, amplitude=7pt, mirror},
      color=black, line width=0.9pt]
  (-0.70,-4.70) -- (4.40,-4.70);
\node[color=black] at (1.85,-5.20) {Encoding};
 
\draw[decorate, decoration={brace, amplitude=7pt, mirror},
      color=black, line width=0.9pt]
  (7.9,-4.70) -- (12.90,-4.70);
\node[color=black] at (10.29,-5.20) {Decoding};
 
\end{tikzpicture}